\documentclass[11pt,a4paper]{article}
\usepackage[hyperref]{acl2020}
\usepackage{times}
\usepackage{latexsym}
\usepackage{multibib}
\usepackage{graphicx}
\usepackage{tabularx}
\usepackage{booktabs}
\usepackage{tabu}
\usepackage{multirow}
\usepackage{fontawesome}
\usepackage{amsmath,calc}
\usepackage{soul}
\usepackage{xspace}
\usepackage{tablefootnote}
\usepackage{array}
\usepackage{ragged2e}
\usepackage{epstopdf}
\usepackage[utf8]{inputenc}
\usepackage[T5,T1]{fontenc}
\usepackage{hyperref}
\usepackage{xcolor}
\usepackage{xstring}

\DeclareTextSymbolDefault{\OHORN}{T5}
\DeclareTextSymbolDefault{\UHORN}{T5}
\DeclareTextSymbolDefault{\ohorn}{T5}
\DeclareTextSymbolDefault{\uhorn}{T5}

\usepackage{microtype}
\aclfinalcopy %
\def\aclpaperid{2183} %

\newcommand{\camembert}{CamemBERT\xspace}

\newcommand{\camembertccnetlarge}{CamemBERT\textsubscript{LARGE}\xspace}
\newcommand{\camembertlarge}{CamemBERT\textsubscript{LARGE}\xspace}

\newcommand{\roberta}{RoBERTa\xspace}
\newcommand{\bert}{BERT\xspace}
\newcommand{\mbert}{mBERT\xspace}

\newcommand{\bertbase}{BERT\textsubscript{BASE}\xspace}
\newcommand{\bertlarge}{BERT\textsubscript{LARGE}\xspace}

\newcommand{\ccnet}{CCNet\xspace}
\newcommand{\oscar}{OSCAR\xspace}

\newcommand{\xlmmlmtlm}{XLM\textsubscript{MLM-TLM}\xspace}
\newcolumntype{P}[1]{>{\RaggedRight\hspace{0pt}}p{#1}}

\usepackage[normalem]{ulem}
\usepackage{soulutf8}
\newcommand\red[1]{\textcolor{red}{{#1}}}

\iftrue %
    \usepackage[final]{proofing}
  \renewcommand\hl[1]{{#1}}  %
   {\draftnote{\red{#2}}}
   \newcommand\redHL[1]{}
  \aclfinalcopy %

  \newcommand{\Djame}[1]{}

\else  %

\usepackage[draft]{proofing}

\newcommand{\Djame}[1]{
\textbf{\textcolor{red}{\hl{Djame: #1}}}
}

\renewcommand\red[1]{{\textbf{\textcolor{red}{#1}}}}

\let\oldred\red
\renewcommand\red[1]{{\bf \oldred{{#1}}}}

 \newcommand\redHL[1]{\red{\hl{#1}}}
\let\olddraftnote\draftnote
\renewcommand\draftnote[1]{\olddraftnote{\red{#1}}}

\fi

\title{CamemBERT: a Tasty French Language Model}

\author{Louis Martin$^{* 1,2,3}$\quad Benjamin Muller$^{* 2,3}$\quad Pedro Javier Ortiz Su\'arez$^{* 2,3}$\quad Yoann Dupont$^3$\\ \large\textbf{Laurent Romary$^2$\quad \'Eric Villemonte de la Clergerie$^2$\quad Djam\'e Seddah$^2$\quad Beno\^it Sagot$^2$}\\
  $^{1}$Facebook AI Research, Paris, France \\
  $^{2}$Inria, Paris, France \\
  $^{3}$Sorbonne Universit\'e, Paris, France\\
  \texttt{louismartin@fb.com}\\
  \texttt{\{benjamin.muller, pedro.ortiz, laurent.romary,}\\
  \texttt{eric.de\_la\_clergerie, djame.seddah, benoit.sagot\}@inria.fr}\\ 
  \texttt{yoa.dupont@gmail.com}}

\date{}

\begin{document}

\maketitle

\begin{abstract}

Pretrained language models are now ubiquitous in Natural Language Processing. 
Despite their success, most available models have either been trained on English data or on the concatenation of data in multiple languages. This makes practical use of such models---in all languages except English---very limited. In this paper, we investigate the feasibility of training monolingual Transformer-based language models for other languages, taking French as an example and evaluating our language models on part-of-speech tagging, dependency parsing, named entity recognition and natural language inference tasks. We show that the use of web crawled data is preferable to the use of Wikipedia data. More surprisingly, we show that a relatively small web crawled dataset (4GB) leads to results that are as good as those obtained using larger datasets (130+GB). Our best performing model \camembert reaches or improves the state of the art in all four downstream tasks. %

\end{abstract}

\renewcommand{\thefootnote}{\fnsymbol{footnote}}
\footnotetext[1]{Equal contribution. Order determined alphabetically.}
\renewcommand{\thefootnote}{\arabic{footnote}}

\section{Introduction}

Pretrained word representations have a long history in Natural Language Processing (NLP), from non-contextual \cite{brown1992class,ando2005framework,mikolov2013distributed,pennington2014glove} to contextual word embeddings \cite{peters2018deep,akbik2018contextual}.
Word representations are usually obtained by training language model architectures on large amounts of textual data and then fed as an input to more complex task-specific architectures.
More recently, these specialized architectures have been replaced altogether by large-scale pretrained language models which are {\em fine-tuned} for each application considered.
This shift has resulted in large improvements in performance over a wide range of tasks \cite{devlin2019bert,radford2019language,liu2019roberta,raffel2019exploring}.

These transfer learning methods exhibit clear advantages over more traditional task-specific approaches. In particular, they can be trained in an \emph{unsupervized} manner, thereby taking advantage of the information contained in large amounts of raw text.
Yet they come with implementation challenges, namely the amount of data and computational resources needed for pretraining, which can reach hundreds of gigabytes of text and require hundreds of GPUs \cite{yang2019xlnet,liu2019roberta}.
This has limited the availability of these state-of-the-art models to the English language, at least in the monolingual setting.
This is particularly inconvenient as it hinders their practical use in NLP systems. It also prevents us from investigating their language modelling capacity, for instance in the case of morphologically rich languages.

Although multilingual models give remarkable results, they are often larger, and their results, as we will observe for French, can lag behind their monolingual counterparts for high-resource languages. %

In order to reproduce and validate results that have so far only been obtained for English, we take advantage of the newly available multilingual corpora OSCAR \cite{ortiz2019asynchronous} to train a monolingual language model for French, dubbed \camembert. We also train alternative versions of \camembert on different smaller corpora with different levels of homogeneity in genre and style in order to assess the impact of these parameters on downstream task performance.
\camembert uses the \roberta architecture \cite{liu2019roberta}, an improved variant of the high-performing and widely used \bert architecture \cite{devlin2019bert}.

We evaluate our model on four different downstream tasks for French: part-of-speech (POS) tagging, dependency parsing, named entity recognition (NER) and natural language inference (NLI).
\camembert improves on the state of the art in all four tasks compared to previous monolingual and multilingual approaches including \mbert, XLM and XLM-R, which confirms the effectiveness of large pretrained language models for French.

Our contributions can be summarized as follows:
\begin{itemize}
    \item First to release a monolingual \roberta model for the French language using recently introduced large-scale open source corpora from the Oscar collection and first outside the original \bert authors to release such a large model for an other language than English. This model is made available publicly under an MIT open-source license\footnote{Released at:
    \mbox{\url{https://camembert-model.fr}}}.
    \item We achieve state-of-the-art results on four downstream tasks: POS tagging, dependency parsing, NER and NLI, confirming the effectiveness of \bert-based language models for French.
    \item We demonstrate that small and diverse training sets can achieve similar performance to large-scale corpora, by analysing the importance of the pretraining corpus in terms of size and domain.%
\end{itemize}

\section{Previous work}
\label{relatedwork}
\subsection{Contextual Language Models}
\paragraph{From non-contextual to contextual word embeddings}
The first neural word vector representations were non-contextualized word embeddings, most notably
 word2vec \cite{mikolov2013distributed}, GloVe \cite{pennington2014glove} and fastText \cite{mikolov2018advances}, which were designed to be used as input to task-specific neural architectures.
 Contextualized word representations such as ELMo \cite{peters2018deep} and flair \cite{akbik2018contextual}, improved the representational power of word embeddings by taking context into account. Among other reasons, they improved the performance of models on many tasks by handling words polysemy.
 This paved the way for larger contextualized models that replaced downstream architectures altogether in most tasks. Trained with language modeling objectives, these approaches range from LSTM-based architectures such as \cite{dai2015semisupervised}, to the successful transformer-based architectures such
 as GPT2 \cite{radford2019language}, \bert \cite{devlin2019bert}, \roberta \cite{liu2019roberta} and more recently ALBERT \cite{lan2019albert} and T5 \cite{raffel2019exploring}.

\paragraph{Non-English contextualized models}
\label{contextualmodelsforotherlanguages}
Following the success of large pretrained language models, they were extended to the multilingual setting with multilingual \bert (hereafter \mbert) \cite{devlin2018mbert}, a single multilingual model for 104 different languages trained on Wikipedia data, and later XLM \cite{lample2019cross}, which significantly improved unsupervized machine translation.
More recently XLM-R \cite{conneau2019xlmr}, extended XLM by training on 2.5TB of data and outperformed previous scores on multilingual benchmarks. They show that multilingual models can obtain results competitive with monolingual models by leveraging higher quality data from other languages on specific downstream tasks.

A few non-English monolingual models have been released: ELMo models for Japanese, Portuguese, German and Basque\footnote{\url{https://allennlp.org/elmo}} and BERT for Simplified and Traditional Chinese \cite{devlin2018mbert} and German \cite{chan2019german}.

However, to the best of our knowledge, no particular effort has been made toward training models for languages other than English at a scale similar to the latest English models (e.g.~\roberta trained on more than 100GB of data).

\paragraph{BERT and RoBERTa}
Our approach is based on \roberta \cite{liu2019roberta} which itself is based on \bert \cite{devlin2019bert}.
\bert is a multi-layer bidirectional Transformer encoder trained with a masked language modeling (MLM) objective, inspired by the Cloze task \cite{taylor1953cloze}.
It comes in two sizes: the \bertbase architecture and the \bertlarge architecture. The \bertbase architecture is 3 times smaller and therefore faster and easier to use while \bertlarge achieves increased performance on downstream tasks.
\roberta improves the original implementation of \bert by identifying key design choices for better performance, using dynamic masking, removing the next sentence prediction task, training with larger batches, on more data, and for longer.

\section{Downstream evaluation tasks}

In this section, we present the four downstream tasks that we use to evaluate \camembert, namely: Part-Of-Speech (POS) tagging, dependency parsing, Named Entity Recognition (NER) and Natural Language Inference (NLI). We also present the baselines that we will use for comparison.

\paragraph{Tasks} POS tagging is a low-level syntactic task, which consists in assigning to each word its corresponding grammatical category. Dependency parsing consists in predicting the labeled syntactic tree in order to capture the syntactic relations between words.

For both of these tasks we run our experiments using the Universal Dependencies (UD)\footnote{\url{https://universaldependencies.org}} framework and its corresponding UD POS tag set \citep{petrov2011universal} and UD treebank collection \citep{ud22}, which was used for the CoNLL 2018 shared task \citep{seker2018universal}. We perform our evaluations on the four freely available French UD treebanks in UD~v2.2: GSD \citep{mcdonald13}, Sequoia\footnote{\url{https://deep-sequoia.inria.fr}} \citep{candito2012le,candito2014deep}, Spoken \citep{lacheret14,bawden14}\footnote{Speech transcript uncased that includes annotated disfluencies without punctuation}, and ParTUT \cite{sanguinetti2015PartTUT}. A brief overview of the size and content of each treebank can be found in Table \ref{treebanks-tab}.

\begin{table}[ht]
\centering\small
\resizebox{\linewidth}{!}{
\begin{tabular}{lccl}
    \toprule
    Treebank & \#Tokens  & \#Sentences & \multicolumn{1}{l}{Genres} \\
    \midrule
    & & & Blogs, News\\
    \multirow{-2}{*}[1.5pt]{GSD} & \multirow{-2}{*}[1.5pt]{389,363} & \multirow{-2}{*}[1.5pt]{16,342} & Reviews, Wiki \\ \tabucline[\hbox {$\scriptstyle \cdot$}]{-}
    & & & Medical, News \\
    \multirow{-2}{*}[0.7pt]{Sequoia} & \multirow{-2}{*}[0.7pt]{68,615} & \multirow{-2}{*}[0.7pt]{3,099} &  Non-fiction, Wiki \\ \tabucline[\hbox {$\scriptstyle \cdot$}]{-}
    Spoken & 34,972  & 2,786 & Spoken \\ \tabucline[\hbox {$\scriptstyle \cdot$}]{-}
    ParTUT & 27,658 & 1,020 & Legal, News, Wikis \\ \tabucline[\hbox {$\scriptstyle \cdot$}]{-}
    FTB & 350,930 & 27,658 & News \\
    \bottomrule
\end{tabular}
}
\caption{Statistics on the treebanks used in POS tagging, dependency parsing, and NER (FTB).}\label{treebanks-tab}
\end{table}

We also evaluate our model in NER, which is a sequence labeling task predicting which words refer to real-world objects, such as people, locations, artifacts and organisations. We use the French Treebank\footnote{This dataset has only been stored and used on Inria's servers after signing the research-only agreement.} (FTB) \citep{abeille:03} in its 2008 version introduced by \citet{cc-clustering:09short} and with NER annotations by \citet{sagot2012annotation}. The FTB contains more than 11 thousand entity mentions distributed among 7 different entity types. A brief overview of the FTB can also be found in Table \ref{treebanks-tab}.

Finally, we evaluate our model on NLI, using the French part of the XNLI dataset \cite{conneau2018xnli}. NLI consists in predicting whether a hypothesis sentence is entailed, neutral or contradicts a premise sentence. The XNLI dataset is the extension of the Multi-Genre NLI (MultiNLI) corpus \cite{williams2018broad} to 15 languages by translating the validation and test sets manually into each of those languages.
The English training set is machine translated for all languages other than English.
The dataset is composed of 122k train, 2490 development and 5010 test examples for each language.
As usual, NLI performance is evaluated using accuracy.

\paragraph{Baselines}
In dependency parsing and POS-tagging we compare our model with:

\begin{itemize}
    \item \emph{\mbert}: The multilingual cased version of \bert (see Section~\ref{contextualmodelsforotherlanguages}). We fine-tune \mbert on each of the treebanks with an additional layer for POS-tagging and dependency parsing, in the same conditions as our \camembert model.
    \item \emph{\xlmmlmtlm}: A multilingual pretrained language model from \citet{lample2019cross}, which showed better performance than \mbert on NLI. We use the version available in the Hugging's Face transformer library \cite{Wolf2019HuggingFacesTS}; like \mbert, we fine-tune it in the same conditions as our model.
    \item \emph{UDify} \cite{kondratyuk201975}: A multitask and multilingual model based on \mbert, UDify is trained simultaneously on 124 different UD treebanks, creating a single POS tagging and dependency parsing model that works across 75 different languages. We report the scores from \citet{kondratyuk201975} paper.
    \item \emph{UDPipe Future} \citep{straka2018udpipe}: An LSTM-based model ranked 3\textsuperscript{rd} in dependency parsing and 6\textsuperscript{th} in POS tagging at the CoNLL~2018 shared task \citep{seker2018universal}. We report the scores from \citet{kondratyuk201975} paper.
    \item \emph{UDPipe Future + \mbert + Flair} \citep{straka2019evaluating}: The original UDPipe Future implementation using \mbert and Flair as feature-based contextualized word embeddings. We report the scores from \citet{straka2019evaluating} paper.
\end{itemize}

In French, no extensive work has been done on NER due to the limited availability of annotated corpora. Thus we compare our model with the only recent available baselines set by \citet{dupont2018exploration}, who trained both CRF \citep{lafferty2001conditional} and BiLSTM-CRF \citep{lample2016neural} architectures on the FTB and enhanced them using heuristics and pretrained word embeddings. Additionally, as for POS and dependency parsing, we compare our model to a fine-tuned version of \mbert for the NER task.

For XNLI, we provide the scores of \mbert which has been reported for French by \citet{wu2019beto}.
We report scores from \xlmmlmtlm (described above), the best model from \citet{lample2019cross}. %
We also report the results of \mbox{XLM-R} \cite{conneau2019xlmr}.

\section{\camembert: a French Language Model}\label{sec:Camembert}
In this section, we describe the pretraining data, architecture, training objective and optimisation setup we use for \camembert.

\subsection{Training data}
Pretrained language models benefits from being trained on large datasets \cite{devlin2018mbert,liu2019roberta,raffel2019exploring}. %
We therefore use the French part of the OSCAR corpus \cite{ortiz2019asynchronous}, a pre-filtered and pre-classified version of Common Crawl.\footnote{\url{https://commoncrawl.org/about/}} %

\paragraph{OSCAR} is a set of monolingual corpora extracted from Common Crawl snapshots{}. It follows the same approach as \cite{grave2018learning} by using a language classification model based on the fastText linear classifier \cite{grave2017bag,joulin2016fasttext} pretrained on Wikipedia, Tatoeba and SETimes, which supports 176 languages. No other filtering is done. We use a non-shuffled version of the French data, which amounts to 138GB of raw text and 32.7B tokens after subword tokenization.%

\subsection{Pre-processing}
We segment the input text data into subword units using SentencePiece \cite{kudo2018sentencepiece}.
SentencePiece is an extension of Byte-Pair encoding (BPE) \cite{sennrich2016neural} and WordPiece \cite{kudo2018subword} that does not require pre-tokenization (at the word or token level), thus removing the need for language-specific tokenisers.
We use a vocabulary size of 32k subword tokens. These subwords are learned on $10^7$ sentences sampled randomly from the pretraining dataset.
We do not use subword regularisation (i.e.~sampling from multiple possible segmentations) for the sake of simplicity.

\subsection{Language Modeling}

\paragraph{Transformer}
Similar to \roberta and \bert, \camembert is a multi-layer bidirectional Transformer \cite{vaswani2017attention}. Given the widespread usage of Transformers, we do not describe them here and refer the reader to \citep{vaswani2017attention}.
\camembert uses the original architectures of \bertbase (12 layers, 768 hidden dimensions, 12 attention heads, 110M parameters) and \bertlarge (24 layers, 1024 hidden dimensions, 12 attention heads, 110M parameters).
\camembert is very similar to \roberta, the main difference being the use of whole-word masking and the usage of SentencePiece tokenization \cite{kudo2018sentencepiece} instead of WordPiece \cite{schuster2012japanese}.

\paragraph{Pretraining Objective}
We train our model on the Masked Language Modeling (MLM) task.
Given an input text sequence composed of $N$ tokens $x_1, ..., x_N$, we select 15\% of tokens for possible replacement. Among those selected tokens, 80\% are replaced with the special \texttt{<MASK>} token, 10\% are left unchanged and 10\% are replaced by a random token. The model is then trained to predict the initial masked tokens using cross-entropy loss.

Following the \roberta approach, we dynamically mask tokens instead of fixing them statically for the whole dataset during preprocessing. This improves variability and makes the model more robust when training for multiple epochs.

Since we use SentencePiece to tokenize our corpus, the input tokens to the model are a mix of whole words and subwords.
An upgraded version of \bert\footnote{\url{https://github.com/google-research/bert/blob/master/README.md}} and \citet{joshi2019spanbert} have shown that masking whole words instead of individual subwords leads to improved performance.
Whole-word Masking (WWM) makes the training task more difficult because the model has to predict a whole word rather than predicting only part of the word given the rest.
We train our models using WWM by using whitespaces in the initial untokenized text as word delimiters.

WWM is implemented by first randomly sampling 15\% of the words in the sequence and then considering all subword tokens in each of this 15\% for candidate replacement. This amounts to a proportion of selected tokens that is close to the original 15\%.
These tokens are then either replaced by \texttt{<MASK>} tokens (80\%), left unchanged (10\%) or replaced by a random token.

Subsequent work has shown that the next sentence prediction (NSP) task originally used in \bert does not improve downstream task performance \cite{lample2019cross,liu2019roberta}, thus we also remove it.

\paragraph{Optimisation}
Following \citep{liu2019roberta}, we optimize the model using Adam \cite{kingma2014adam} ($\beta_1 = 0.9$, $\beta_2 = 0.98$) for 100k steps with large batch sizes of 8192 sequences, each sequence containing at most 512 tokens.
We enforce each sequence to only contain complete paragraphs (which correspond to lines in the our pretraining dataset).

\paragraph{Pretraining}
We use the \roberta implementation in the fairseq library \cite{ott2019fairseq}.
Our learning rate is warmed up for 10k steps up to a peak value of $0.0007$ instead of the original $0.0001$ given our large batch size, and then fades to zero with polynomial decay.
Unless otherwise specified, our models use the BASE architecture, and are pretrained for 100k backpropagation steps on 256 Nvidia V100 GPUs (32GB each) for a day.
We do not train our models for longer due to practical considerations, even though the performance still seemed to be increasing.

\subsection{Using \camembert for downstream tasks}
We use the pretrained \camembert in two ways. In the first one, which we refer to as \textit{fine-tuning}, we fine-tune the model on a specific task in an end-to-end manner. In the second one, referred to as \textit{feature-based embeddings} or simply \textit{embeddings}, we extract frozen contextual embedding vectors from \camembert.
These two complementary approaches shed light on the quality of the pretrained hidden representations captured by \camembert.

\paragraph{Fine-tuning}
For each task, we append the relevant predictive layer on top of \camembert's  architecture. Following the work done on \bert \cite{devlin2019bert}, for sequence tagging and sequence labeling we append a linear layer that respectively takes as input the last hidden representation of the \texttt{<s>} special token and the last hidden representation of the first subword token of each word.
For dependency parsing, we plug a bi-affine graph predictor head as inspired by \citet{dozat2017deep}. We refer the reader to this article for more details on this module.
We fine-tune on XNLI by adding a classification head composed of one hidden layer with a non-linearity and one linear projection layer, with input dropout for both.

We fine-tune \camembert independently for each task and each dataset. We optimize the model using the Adam optimiser \cite{kingma2014adam} with a fixed learning rate. We run a grid search on a combination of learning rates and batch sizes. We select the best model on the validation set out of the 30 first epochs.
For NLI we use the default hyper-parameters provided by the authors of RoBERTa on the MNLI task.\footnote{More details at \url{https://github.com/pytorch/fairseq/blob/master/examples/roberta/README.glue.md}.}
Although this might have pushed the performances even further, we do not apply any regularisation techniques such as weight decay, learning rate warm-up or discriminative fine-tuning, except for NLI. We show that fine-tuning \camembert in a straightforward manner leads to state-of-the-art results on all tasks and outperforms the existing \bert-based models in all cases. 
The POS tagging, dependency parsing, and NER experiments are run using Hugging Face's Transformer library extended to support \camembert and dependency parsing \cite{Wolf2019HuggingFacesTS}. 
The NLI experiments use the fairseq library following the \roberta implementation.

\paragraph{Embeddings}

Following \citet{strakova2019neural} and \citet{straka2019evaluating} for \mbert and the English BERT, we make use of \camembert in a feature-based embeddings setting.
In order to obtain a representation for a given token, we first compute the average of each sub-word’s representations in the last four layers of the Transformer, and then average the resulting sub-word vectors.

We evaluate \camembert in the embeddings setting for POS tagging, dependency parsing and NER; using the open-source implementations of \newcite{straka2019evaluating} and \newcite{strakova2019neural}.\footnote{UDPipe Future is available at \url{https://github.com/CoNLL-UD-2018/UDPipe-Future}, and the code for nested NER is available at \url{https://github.com/ufal/acl2019_nested_ner}.}

\section{Evaluation of \camembert}

In this section, we measure the performance of our models by evaluating them on the four aforementioned tasks: POS tagging, dependency parsing, NER and NLI.

\begin{table*}[ht]
\small\centering
\resizebox{\linewidth}{!}{
\begin{tabu}{ l  c  c @{\hspace{0.35cm}}  @{\hspace{0.35cm}} c  c @{\hspace{0.35cm}}  @{\hspace{0.35cm}} c  c  @{\hspace{0.35cm}}  @{\hspace{0.35cm}} c  c }
	\toprule
	& \multicolumn{2}{c @{\hspace{0.5cm}}}{\textsc{GSD}} & \multicolumn{2}{c @{\hspace{0.7cm}}}{\textsc{Sequoia}} & \multicolumn{2}{c @{\hspace{0.7cm}}}{\textsc{Spoken}} & \multicolumn{2}{c @{\hspace{0.35cm}}}{\textsc{ParTUT}} \\
	\cmidrule(l{2pt}r{0.4cm}){2-3}\cmidrule(l{-0.2cm}r{0.4cm}){4-5}\cmidrule(l{-0.2cm}r{0.4cm}){6-7}\cmidrule(l{-0.2cm}r{2pt}){8-9}
	\multirow{-2}{*}[1pt]{\textsc{Model}} & \textsc{UPOS}     & \textsc{LAS}      & \textsc{UPOS}  & \textsc{LAS}      & \textsc{UPOS}     & \textsc{LAS}      & \textsc{UPOS}     & \textsc{LAS}      \\
	\midrule
	\mbert  (fine-tuned)                                  & 97.48             & 89.73             & 98.41          & 91.24 & 96.02             & 78.63             & 97.35 & 91.37 \\
	\xlmmlmtlm (fine-tuned)                         &        98.13      & 90.03                & 98.51          & 91.62             & 96.18             & 80.89            & 97.39             & 89.43             \\ %
	UDify \cite{kondratyuk201975}                   & 97.83 & \underline{91.45} & 97.89          & 90.05             & 96.23 & 80.01 & 96.12             & 88.06             \\
	UDPipe Future \cite{straka2018udpipe}                        & 97.63             & 88.06             & 98.79          & 90.73             & 95.91             & 77.53             & 96.93             & 89.63             \\
	\: + mBERT + Flair  (emb.) \cite{straka2019evaluating}                   & \underline{97.98}             & 90.31             & \textbf{99.32} & 93.81             & \textbf{97.23}    & \underline{81.40}             & \underline{97.64}             & \underline{92.47}             \\
	\tabucline[\hbox {$\scriptstyle \cdot$}]{-}
	\camembert (fine-tuned)                           & \textbf{98.18}    & \textbf{92.57}    & \underline{99.29}          & \textbf{94.20}    & 96.99             & 81.37             & \textbf{97.65}    & \textbf{93.43}             \\ %
	UDPipe Future \mbox{+ \camembert} (embeddings)    & 97.96             & 90.57             & 99.25          & \underline{93.89}             & \underline{97.09}             & \textbf{81.81}             & 97.50             & 92.32             \\
	\bottomrule
\end{tabu}
}
\caption{\textbf{POS} and \textbf{dependency parsing} scores on 4 French treebanks, reported on test sets assuming gold tokenization and segmentation (best model selected on validation out of 4). Best scores in bold, second best underlined.}%
\label{tab:pos_and_dp_results}
\end{table*}

\begin{table}[ht]
    \centering \small
    \scalebox{0.9}{
        \begin{tabu}{lc}
            \toprule
            Model   & F1 \\
            \midrule
            SEM (CRF) \cite{dupont2018exploration} & 85.02  \\
            LSTM-CRF \cite{dupont2018exploration} & 85.57  \\
            \mbert (fine-tuned)  & 87.35  \\
            \tabucline[\hbox {$\scriptstyle \cdot$}]{-}
            \camembert (fine-tuned) &  \underline{89.08} \\
            LSTM+CRF+\camembert (embeddings)  & \textbf{89.55} \\
            \bottomrule
        \end{tabu}
        }
    \caption{\textbf{NER} scores on the FTB (best model selected on validation out of 4). Best scores in bold, second best underlined. 
    \label{table:ner_ablation}}
\end{table}

\begin{table}[ht]
    \centering\small
    \scalebox{0.9}{
        \begin{tabu}{lcc}
            \toprule
            Model & Acc. & \#Params  \\
            \midrule
            \mbert \cite{devlin2019bert} & 76.9 & 175M \\
            \xlmmlmtlm \cite{lample2019cross} & \underline{80.2} & 250M \\
            XLM-R\textsubscript{BASE} \cite{conneau2019xlmr} & 80.1 & 270M \\
            \tabucline[\hbox {$\scriptstyle \cdot$}]{-}
            \camembert (fine-tuned) & \textbf{82.5} & 110M \\ 
            \midrule
            \multicolumn{3}{c}{\em Supplement: LARGE models}\\
            XLM-R\textsubscript{LARGE} \cite{conneau2019xlmr} & \underline{85.2} & 550M \\
            \tabucline[\hbox {$\scriptstyle \cdot$}]{-}
            \camembertccnetlarge (fine-tuned) & \textbf{85.7} & 335M \\ 
            \bottomrule
        \end{tabu}
    }
    \caption{\textbf{NLI} accuracy on the French XNLI test set (best model selected on validation out of 10). Best scores in bold, second best underlined.\label{table:xnli}}
\end{table}

\paragraph{POS tagging and dependency parsing}
For POS tagging and dependency parsing, we compare \camembert with other models in the two settings: \textit{fine-tuning} and as \textit{feature-based embeddings}.
We report the results in Table~\ref{tab:pos_and_dp_results}.

\camembert reaches state-of-the-art scores on all treebanks and metrics in both scenarios. The two approaches achieve similar scores, with a slight advantage for the fine-tuned version of \camembert, thus questioning the need for complex task-specific architectures such as UDPipe Future.

Despite a much simpler optimisation process and no task specific architecture, fine-tuning \camembert outperforms UDify on all treebanks and sometimes by a large margin (e.g. +4.15\% LAS on Sequoia and +5.37 LAS on ParTUT).
\camembert also reaches better performance  than other multilingual pretrained models such as \mbert and \xlmmlmtlm on all treebanks.%

\camembert achieves overall slightly better results than the previous state-of-the-art and task-specific architecture UDPipe Future+\mbert+Flair, except for POS tagging on Sequoia and POS tagging on Spoken, where \camembert lags by 0.03\% and 0.14\% UPOS respectively.
UDPipe Future+\mbert+Flair uses the contextualized string embeddings Flair \citep{akbik2018contextual}, which are in fact pretrained contextualized character-level word embeddings specifically designed to handle misspelled words as well as subword structures such as prefixes and suffixes. This design choice might explain the difference in score for POS tagging with CamemBERT, especially for the Spoken treebank where words are not capitalized, a factor that might pose a problem for CamemBERT which was trained on capitalized data, but that might be properly handle by Flair on the UDPipe Future+\mbert+Flair model.

\paragraph{Named-Entity Recognition}
For NER, we similarly evaluate \camembert in the fine-tuning setting and as input embeddings to the task specific architecture LSTM+CRF. We report these scores in Table~\ref{table:ner_ablation}.

In both scenarios, \camembert achieves higher F1 scores than the traditional CRF-based architectures, both non-neural and neural, and than fine-tuned multilingual BERT models.\footnote{\xlmmlmtlm is a lower-case model. Case is crucial for NER, therefore we do not report its low performance (84.37\%)}

Using \camembert as embeddings to the traditional LSTM+CRF architecture gives slightly higher scores than by fine-tuning the model (89.08 vs.~89.55).
This demonstrates that although \camembert can be used successfully without any task-specific architecture, it can still produce high quality contextualized embeddings that might be useful in scenarios where powerful downstream architectures exist.

\paragraph{Natural Language Inference}
On the XNLI benchmark, we compare \camembert to previous state-of-the-art multilingual models in the fine-tuning setting. In addition to the standard \camembert model with a BASE architecture, we train another model with the LARGE architecture, referred to as \camembertccnetlarge, for a fair comparison with XLM-R\textsubscript{LARGE}.
This model is trained with the \ccnet corpus, described in Sec.~\ref{sec:origin_and_size}, for 100k steps.\footnote{We train our LARGE model with the \ccnet corpus for practical reasons. Given that BASE models reach similar performance when using \oscar or \ccnet as pretraining corpus (Appendix Table~\ref{tab:ablation}), we expect an \oscar LARGE model to reach comparable scores.} We expect that training the model for longer would yield even better performance.

\camembert reaches higher accuracy than its BASE counterparts reaching +5.6\% over \mbert, +2.3 over \xlmmlmtlm, and +2.4 over XLM-R\textsubscript{BASE}. \camembert also uses as few as half as many parameters (110M vs. 270M for XLM-R\textsubscript{BASE}).

\camembertccnetlarge achieves a state-of-the-art accuracy of 85.7\% on the XNLI benchmark, as opposed to 85.2, for the recent XLM-R\textsubscript{LARGE}.

\camembert uses fewer parameters than multilingual models, mostly because of its smaller vocabulary size (e.g. 32k vs. 250k for XLM-R).
Two elements might explain the better performance of \camembert over XLM-R.
Even though XLM-R was trained on an impressive amount of data (2.5TB), only 57GB of this data is in French, whereas we used 138GB of French data.
Additionally XLM-R also handles 100 languages, and the authors show that when reducing the number of languages to 7, they can reach 82.5\% accuracy for French XNLI with their BASE architecture.

\paragraph{Summary of \camembert's results}
\camembert improves the state of the art for the 4 downstream tasks considered, thereby confirming on French the usefulness of Transformer-based models. We obtain these results when using \camembert as a fine-tuned model or when used as contextual embeddings with task-specific architectures.
This questions the need for more complex downstream architectures, similar to what was shown for English \cite{devlin2019bert}.
Additionally, this suggests that \camembert is also able to produce high-quality representations out-of-the-box without further tuning.

\begin{table*}[ht]
	\small\centering
	\resizebox{\textwidth}{!}{
		\tabulinesep =_1pt^1pt
		\begin{tabu}{ l l @{\hspace{0.7cm}}  c  c  @{\hspace{0.7cm}} c  c  @{\hspace{0.7cm}} c  c @{\hspace{0.7cm}} c  c @{\hspace{0.7cm}} c c @{\hspace{0.7cm}} c @{\hspace{0.7cm}} c @{\hspace{0.7cm}}}
			\toprule
			& & \multicolumn{2}{c @{\hspace{0.5cm}}}{\textsc{GSD}} & \multicolumn{2}{c @{\hspace{0.7cm}}}{\textsc{Sequoia}} & \multicolumn{2}{c @{\hspace{0.7cm}}}{\textsc{Spoken}} & \multicolumn{2}{c @{\hspace{0.7cm}}}{\textsc{ParTUT}} & \multicolumn{2}{c @{\hspace{0.7cm}}}{\textsc{\textbf{Average}}} & NER & NLI \\
			\cmidrule(l{2pt}r{0.4cm}){3-4}\cmidrule(l{-0.2cm}r{0.4cm}){5-6}\cmidrule(l{-0.2cm}r{0.4cm}){7-8}\cmidrule(l{-0.2cm}r{0.4cm}){9-10}\cmidrule(l{-0.2cm}r{0.4cm}){11-12}\cmidrule(l{-0.2cm}r{0.4cm}){13-13}\cmidrule(l{-0.2cm}r{0.4cm}){14-14}
			\multirow{-2}{*}[2pt]{\textsc{Dataset}} & \multirow{-2}{*}[2pt]{\textsc{Size}} & \textsc{UPOS}  & \textsc{LAS}      & \textsc{UPOS}     & \textsc{LAS}      & \textsc{UPOS}     & \textsc{LAS}      & \textsc{UPOS}  & \textsc{LAS}   & \textsc{UPOS} & \textsc{LAS} &  \textsc{F1}  & \textsc{Acc.}     \\
			\midrule
			
			\multicolumn{10}{l}{\hspace*{6mm}\em Fine-tuning}\\[0.5mm]
			Wiki                               & 4GB                                  & 98.28          & 93.04             & 98.74             & 92.71             & 96.61             & 79.61             & 96.20          & 89.67             &   97.45  &  88.75  & 89.86             & 78.32             \\ %
			\ccnet                                  & 4GB                                  & 98.34          & 93.43             & 98.95             & 93.67             & 96.92             & \textbf{82.09}             & 96.50          & \textbf{90.98}             &  97.67 & \textbf{90.04}  & 90.46             & \textbf{82.06} \\
			\oscar                                  & 4GB                                  & \underline{98.35}          & \underline{93.55} & \underline{98.97}             & \underline{93.70}             & \underline{96.94}             & \underline{81.97}             & \underline{96.58}          & 90.28             &  \underline{97.71} & 89.87  & \underline{90.65}             & \underline{81.88}             \\
			\tabucline[\hbox{$\scriptstyle \cdot$}]{-}
			\oscar                                  & 138GB                                & \textbf{98.39} & \textbf{93.80}    & \textbf{98.99}             & \textbf{94.00}             & \textbf{97.17}             & 81.18             & \textbf{96.63}          & \underline{90.56} &  \textbf{97.79} &  \underline{89.88} &   \textbf{91.55}             & 81.55             \\
			\midrule 
			\multicolumn{11}{l}{\hspace*{6mm}\em Embeddings (with UDPipe Future (tagging, parsing) or LSTM+CRF (NER))} \\[0.5mm] 
			Wiki                               & 4GB                                  & 98.09          & 92.31             & 98.74             & 93.55             & 96.24             & 78.91             & 95.78          & 89.79             &  97.21  &  88.64  & 91.23             & -                 \\
			\ccnet                                  & 4GB                                  & \textbf{98.22}          & \textbf{92.93}             & \underline{99.12} & \underline{94.65} & 97.17             & \textbf{82.61}    & \underline{\textbf{96.74}} & \underline{89.95}             &  \underline{97.81}  &  \underline{90.04}  & \textbf{92.30}    & -                 \\
			\oscar                                  & 4GB                                  & \underline{98.21}          & \underline{92.77}             & \underline{99.12} & \textbf{94.92}    & \underline{97.20} & \underline{82.47} & \underline{\textbf{96.74}} & \textbf{90.05}             &  \textbf{97.82}  &  \textbf{90.05}  & \underline{91.90} & -                 \\
			\tabucline[\hbox{$\scriptstyle \cdot$}]{-}
			\oscar                                  & 138GB                                & 98.18          & \underline{92.77}             & \textbf{99.14}    & 94.24             & \textbf{97.26}    & 82.44             & 96.52          & 89.89             &  97.77  &  89.84  & 91.83             & -                 \\
			
			\bottomrule
		\end{tabu}
	}
	\caption{Results on the four tasks using language models pre-trained on data sets of varying homogeneity and size, reported on validation sets (average of 4 runs for POS tagging, parsing and NER, average of 10 runs for NLI).}
	
	\label{tab:ablation_data_size}
\end{table*}

\section{Impact of corpus origin and size}
\label{sec:origin_and_size}

In this section we investigate the influence of the homogeneity and size of the pretraining corpus on downstream task performance. With this aim, we train alternative version of \camembert by varying the pretraining datasets. For this experiment, we fix the number of pretraining steps to 100k, and allow the number of epochs to vary accordingly (more epochs for smaller dataset sizes). All models use the BASE architecture.

In order to investigate the need for homogeneous clean data versus more diverse and possibly noisier data, we use alternative sources of pretraining data in addition to \oscar:
\begin{itemize}
    \item \textbf{Wikipedia}, which is homogeneous in terms of genre and style. We use the official 2019 French Wikipedia dumps\footnote{ \url{https://dumps.wikimedia.org/backup-index.html}.}. We remove HTML tags and tables using Giuseppe Attardi's  \emph{WikiExtractor}.\footnote{ \url{https://github.com/attardi/wikiextractor}.}
    \item \textbf{\ccnet} \cite{wenzek2019ccnet}, a dataset extracted from Common Crawl with a different filtering process than for \oscar. It was built using a language model trained on Wikipedia, in order to filter out bad quality texts such as code or tables.\footnote{We use the \textsc{head} split, which corresponds to the top 33\% of documents in terms of filtering perplexity.} As this filtering step biases the noisy data from Common Crawl to more Wikipedia-like text, we expect \ccnet to act as a middle ground between the unfiltered ``noisy'' \oscar dataset, and the ``clean'' Wikipedia dataset. As a result of the different filtering processes, \ccnet contains longer documents on average compared to \oscar with smaller---and often noisier---documents weeded out.
\end{itemize}
Table~\ref{table:corpora_statistics} summarizes statistics of these different corpora.

\begin{table}[ht]
\centering\small
\resizebox{\linewidth}{!}{
    \begin{tabular}{lcccccc}
    \toprule
   Corpus & Size & \#tokens & \#docs & \multicolumn{3}{c}{Tokens/doc}  \\
    &  &  &  & \multicolumn{3}{c}{Percentiles:}  \\
    &  &  &  & 5\% & 50\% & 95\% \\
    \midrule
    Wikipedia & 4GB & 990M & 1.4M & 102 & 363 & 2530  \\
    CCNet & 135GB & 31.9B & 33.1M & 128 & 414 & 2869 \\
    OSCAR & 138GB & 32.7B & 59.4M & 28 & 201 & 1946 \\
    \bottomrule
\end{tabular}
}
\caption{Statistics on the pretraining datasets used.}
\label{table:corpora_statistics}
\end{table}

In order to make the comparison between these three sources of pretraining data, we randomly sample 4GB of text (at the document level) from \oscar and \ccnet, thereby creating samples of both Common-Crawl-based corpora of the same size as the French Wikipedia. These smaller 4GB samples also provides us a way to investigate the impact of pretraining data size. Downstream task performance for our alternative versions of \camembert are provided in Table~\ref{tab:ablation_data_size}.
The upper section reports scores in the fine-tuning setting while the lower section reports scores for the embeddings.

\subsection{Common Crawl vs.~Wikipedia?}
\label{subsec:homogeneityimpact}

Table~\ref{tab:ablation_data_size} clearly shows that models trained on the 4GB versions of \oscar and \ccnet (Common Crawl) perform consistently better than the the one trained on the French Wikipedia. This is true both in the fine-tuning and embeddings setting. Unsurprisingly, the gap is larger on tasks involving texts whose genre and style are more divergent from those of Wikipedia, such as tagging and parsing on the Spoken treebank.
The performance gap is also very large on the XNLI task, probably as a consequence of the larger diversity of Common-Crawl-based corpora in terms of genres and topics. XNLI is indeed based on multiNLI which covers a range of genres of spoken and written text.

The downstream task performances of the models trained on the 4GB version of \ccnet and \oscar are much more similar.\footnote{We provide the results of a model trained on the whole \ccnet corpus in the Appendix. The conclusions are similar when comparing models trained on the full corpora: downstream results are similar when using \oscar or \ccnet.}

\subsection{How much data do you need?}
\label{subsec:sizeimpact}

An unexpected outcome of our experiments is that the model trained ``only'' on the 4GB sample of \oscar performs similarly  to the standard \camembert trained on the whole 138GB \oscar.
The only task with a large performance gap is NER, where  ``138GB'' models are better by 0.9 F1 points. This could be due to the higher number of named entities present in the larger corpora, which is beneficial for this task. On the contrary, other tasks don't seem to gain from the additional data.

In other words, when trained on corpora such as \oscar and \ccnet, which are heterogeneous in terms of genre and style, 4GB of uncompressed text is large enough as pretraining corpus to reach state-of-the-art results with the BASE architecure, better than those obtained with \mbert (pretrained on 60GB of text).\footnote{The OSCAR-4GB model gets slightly better XNLI accuracy than the full OSCAR-138GB model (81.88 vs. 81.55). This might be due to the random seed used for pretraining, as each model is pretrained only once.} This calls into question the need to use a very large corpus such as \oscar or \ccnet when training a monolingual Transformer-based language model such as BERT or \roberta.
Not only does this mean that the computational (and therefore environmental) cost of training a state-of-the-art language model can be reduced, but it also means that \camembert-like models can be trained for all languages for which a Common-Crawl-based corpus of 4GB or more can be created. \oscar is available in 166 languages, and provides such a corpus for 38 languages. Moreover, it is possible that slightly smaller corpora (e.g.~down to 1GB) could also prove sufficient to train high-performing language models.
We obtained our results with BASE architectures. Further research is needed to confirm the validity of our findings on larger architectures and other more complex natural language understanding tasks.
However, even with a BASE architecture and 4GB of training data, the validation loss is still decreasing beyond 100k steps (and 400 epochs). This suggests that we are still under-fitting the 4GB pretraining dataset, training longer might increase downstream performance.

\section{Discussion}

Since the pre-publication of this work \cite{martinetal2019Camembert},
many monolingual language models have appeared, e.g. 
\cite{le_at_al2019flaubert,FinBert2019,RobBERT2020}, for as much as 30
languages \cite{horvyteam:2020:arxiv}. In almost all tested
configurations they displayed better results than multilingual
language models such as \mbert \cite{pires_et_al2019multilingual}. 
Interestingly, \newcite{le_at_al2019flaubert} showed that using their
FlauBert, a RoBERTa-based language model for French, which was trained on less
but more edited data, in conjunction to \camembert in an ensemble
system could improve the performance of a parsing model and  establish
a new state-of-the-art in constituency parsing of French, highlighting thus
the complementarity of both models.\footnote{We refer the reader to
\cite{le_at_al2019flaubert} for a comprehensive benchmark and details
therein.}
As it was the case for English when \bert was first released, the
availability of similar scale language models for French enabled
interesting applications, such as large scale anonymization of legal
texts, where \camembert-based models established a new
state-of-the-art on this task \cite{bennesty:2019:anonym}, or the first
large question answering experiments on a French Squad data set that
was released very recently \cite{fquad:2020:arXiv} where the authors matched human performance using \camembertlarge.
Being the first pre-trained language model that used the open-source Common Crawl Oscar
corpus and given its impact on the community, 
\camembert paved the way for many works on monolingual language
models that followed. Furthermore, the availability of all its
training data favors reproducibility and is a step towards better
understanding such models.
In that spirit, we make the models used in our experiments available via our website and via the {\tt huggingface} and {\tt fairseq} APIs, in addition to the base \camembert model.

\section{Conclusion}
In this work, we investigated the feasibility of training a
Transformer-based language model for languages other than English.
Using French as an example, we trained \camembert, a language model
based on \roberta.  
We evaluated \camembert on four downstream tasks 
(part-of-speech tagging, dependency parsing, named entity recognition
and natural language inference) in which our best model reached or improved the state of the art in all tasks
considered, even when compared to strong multilingual models such as
\mbert, XLM and XLM-R, while also having fewer parameters.  

Our experiments demonstrate that using web crawled
data with high variability is preferable to using Wikipedia-based data.  In addition we
showed that our models could reach surprisingly high performances with
as low as 4GB of pretraining data, questioning thus the need for large
scale pretraining corpora.  This shows that state-of-the-art
Transformer-based language models can be trained on languages with far
fewer resources than English, whenever a few gigabytes of data are
available. This paves the way for the rise of monolingual contextual
pre-trained language-models for under-resourced languages.  The
question of knowing whether pretraining on small domain specific
content will be a better option than transfer learning techniques such
as fine-tuning remains open and we leave it for future work.

Pretrained on pure open-source corpora, \camembert is freely
available and distributed with the MIT license via popular NLP
libraries (\href{https://github.com/pytorch/fairseq}{\tt fairseq} and \href{https://github.com/huggingface/transformers}{\tt huggingface}) as well as on our website
\href{https://camembert-model.fr}{\tt camembert-model.fr}.

\bigskip 

\section*{Acknowledgments}
We want to thank Clémentine Fourrier for her proofreading and insightful comments, and Alix Chagué for her great logo.
This work was partly funded by three French National funded projects granted to Inria and other partners by the Agence Nationale de la Recherche, namely projects PARSITI (ANR-16-CE33-0021), SoSweet (ANR-15-CE38-0011) and BASNUM (ANR-18-CE38-0003), as well as by the last author's chair in the PRAIRIE institute funded by the French national agency ANR as part of the ``Investissements d’avenir'' programme under the reference \mbox{ANR-19-P3IA-0001}.

\newpage 

\bibliographystyle{acl_natbib}
\bibliography{camembert.bib}

\clearpage
\appendix

\section*{Appendix}

\begin{table*}[ht]
\small\centering
\scalebox{0.8}{
\begin{tabular}{ l l c c c  c @{\hspace{0.35cm}}  @{\hspace{0.35cm}} c  c @{\hspace{0.35cm}}  @{\hspace{0.35cm}} c  c  @{\hspace{0.35cm}}  @{\hspace{0.35cm}} c  c @{\hspace{0.35cm}}  @{\hspace{0.35cm}} c @{\hspace{0.35cm}}  @{\hspace{0.35cm}} c }
\toprule
 &&&& \multicolumn{2}{c @{\hspace{0.5cm}}}{\textsc{GSD}} & \multicolumn{2}{c @{\hspace{0.7cm}}}{\textsc{Sequoia}} & \multicolumn{2}{c @{\hspace{0.7cm}}}{\textsc{Spoken}} & \multicolumn{2}{c @{\hspace{0.7cm}}}{\textsc{ParTUT}} & \textsc{NER} & \textsc{NLI} \\
\cmidrule(l{2pt}r{0.4cm}){5-6}\cmidrule(l{-0.2cm}r{0.4cm}){7-8}\cmidrule(l{-0.2cm}r{0.4cm}){9-10}\cmidrule(l{-0.2cm}r{0.4cm}){11-12}\cmidrule(l{-0.2cm}r{0.4cm}){13-13} \cmidrule(l{-0.2cm}r{2pt}){14-14}
 \multirow{-2}{*}[2pt]{\textsc{Dataset}}&\multirow{-2}{*}[2pt]{\textsc{Masking}}&\multirow{-2}{*}[2pt]{\textsc{Arch.}}&\multirow{-2}{*}[2pt]{\textsc{\#Steps}}& \textsc{UPOS} & \textsc{LAS} & \textsc{UPOS} & \textsc{LAS} & \textsc{UPOS} & \textsc{LAS} & \textsc{UPOS} & \textsc{LAS} & \textsc{F1} & \textsc{Acc.} \\
\midrule

\multicolumn{11}{l}{\hspace*{6mm}\em Fine-tuning} \\[0.5mm]
\toprule
OSCAR          & Subword    & \textsc{Base}  & 100k  & \textbf{98.25}    & 92.29                      & \underline{99.25} & 93.70             & 96.95             & 79.96             & \underline{97.73} & \textbf{92.68}             & 89.23             & 81.18 \\ 
OSCAR          & Whole-word & \textsc{Base}  & 100k  & \underline{98.21} & 92.30                      & 99.21             & \underline{94.33} & 96.97             & 80.16             & \textbf{97.78} & 92.65             & 89.11             & 81.92 \\
CCNET          & Subword    & \textsc{Base}  & 100k  & 98.02             & 92.06                      & \textbf{99.26}    & 94.13             & 96.94             & 80.39             & 97.55          & \underline{92.66}             & 89.05             & 81.77 \\ 
CCNET          & Whole-word & \textsc{Base}  & 100k  & 98.03             & \underline{\textbf{92.43}} & 99.18             & 94.26             & \underline{96.98}             & \underline{80.89}             & 97.46          & 92.33             & \underline{89.27}            & 81.92 \\ 
CCNET          & Whole-word & \textsc{Base}  & 500k  & \underline{98.21} & \underline{\textbf{92.43}} & 99.24             & \textbf{94.60}    & 96.69             & \textbf{80.97}             & 97.65          & 92.48             & 89.08             & \underline{83.43} \\ 
CCNET          & Whole-word & \textsc{Large} & 100k  & 98.01             & 91.09                      & 99.23             & 93.65             & \textbf{97.01}             & \underline{80.89}             & 97.41          & 92.59             & \textbf{89.39}             & \textbf{85.29} \\

\midrule
\multicolumn{11}{l}{\hspace*{6mm}\em Embeddings (with UDPipe Future (tagging, parsing) or LSTM+CRF (NER))} \\[0.5mm]
OSCAR & Subword & \textsc{Base} & 100k & \underline{\textbf{98.01}} & 90.64 & \textbf{99.27} & 94.26 & \underline{97.15} & \textbf{82.56} & \textbf{97.70} & \underline{92.70} & \textbf{90.25} & - \\
OSCAR & Whole-word & \textsc{Base} & 100k & 97.97 & 90.44 & \underline{99.23} & 93.93 & 97.08 & 81.74 & 97.50 & 92.28 & 89.48 & - \\
CCNET & Subword & \textsc{Base} & 100k   & 97.87 & \textbf{90.78} & 99.20 & \underline{94.33} & \textbf{97.17} & \underline{82.39} & \underline{97.54} & 92.51 & 89.38 & - \\
CCNET & Whole-word & \textsc{Base} & 100k  & 97.96 & \underline{90.76} & \underline{99.23} & \textbf{94.34} & 97.04 & 82.09 & 97.39 & \textbf{92.82} & \underline{89.85} & - \\
CCNET & Whole-word & \textsc{Base} & 500k  & 97.84 & 90.25 & 99.14 & 93.96 & 97.01 & 82.17 & 97.27 & 92.28 & 89.07 & - \\
CCNET & Whole-word & \textsc{Large} & 100k  & \underline{\textbf{98.01}} & 90.70 & \underline{99.23} & 94.01 & 97.04 & 82.18 & 97.31 & 92.28 & 88.76 & - \\
\bottomrule
\end{tabular}}
\caption{Performance reported on \textbf{Test sets} for all trained models (\textbf{average} over multiple fine-tuning seeds).}
\label{tab:all_results}
\end{table*}

In the appendix, we analyse different design choices of \camembert (Table~\ref{tab:ablation}), namely with respect to the use of whole-word masking, the training dataset, the model size, and the number of training steps in complement with the analyses of the impact of corpus origin an size (Section~\ref{sec:origin_and_size}. In all the ablations, all scores come from at least 4 averaged runs. For POS tagging and dependency parsing, we average the scores on the 4 treebanks.
We also report all averaged test scores of our different models in Table~\ref{tab:all_results}.

\begin{table*}[!htbp]
\centering\small
\scalebox{1}{
\begin{tabular}{lcccc @{\hspace{0.7cm}} cccc}
    \toprule
    \textsc{Dataset} & \textsc{Masking} & \textsc{Arch.} & \#\textsc{Param.} & \#\textsc{Steps}  & \textsc{UPOS} & \textsc{LAS} & \textsc{NER} & \textsc{XNLI}  \\
    \midrule
    \multicolumn{9}{l}{\hspace*{6mm}\em Masking Strategy}\\
    {\color{gray}\oscar} & Subword & {\color{gray}\textsc{Base}} & {\color{gray}110M} & {\color{gray}100K} & 97.78 & 89.80 & \textbf{91.55}  & 81.04\\
    {\color{gray}\oscar} & Whole-word & {\color{gray}\textsc{Base}} & {\color{gray}110M} & {\color{gray}100K} & \textbf{97.79} & \textbf{89.88} & 91.44  & \textbf{81.55}\\
    \midrule
    \multicolumn{9}{l}{\hspace*{6mm}\em Model Size}\\
    {\color{gray}\ccnet} & {\color{gray}Whole-word} & \textsc{Base}  & 110M & {\color{gray}100K} & 97.67 & 89.46 & 90.13 & 82.22\\
    {\color{gray}\ccnet} & {\color{gray} Whole-word} & \textsc{Large} & 335M & {\color{gray} 100k}  & \textbf{97.74} & \textbf{89.82} & \textbf{92.47} & \textbf{85.73} \\
    \midrule
    \multicolumn{9}{l}{\hspace*{6mm}\em Dataset}\\
    \ccnet & {\color{gray} Whole-word} & {\color{gray}\textsc{Base}} & {\color{gray}110M} & {\color{gray}100K} & 97.67 &  89.46 & 90.13 & \textbf{82.22}\\
    \oscar & {\color{gray} Whole-word} & {\color{gray}\textsc{Base}} & {\color{gray}110M} & {\color{gray}100K} & \textbf{97.79} & \textbf{89.88} & \textbf{91.44}  & 81.55\\
    \midrule
    \multicolumn{9}{l}{\hspace*{6mm}\em Number of Steps}\\
    {\color{gray}\ccnet} & {\color{gray} Whole-word} & {\color{gray} \textsc{Base}} & {\color{gray} 110M} & 100k  & \textbf{98.04} & 89.85 & 90.13  & 82.20 \\
    {\color{gray}\ccnet} & {\color{gray} Whole-word} & {\color{gray} \textsc{Base}} & {\color{gray} 110M} & 500k & 97.95 & \textbf{90.12} & 91.30 & \textbf{83.04} \\
    \bottomrule
    \end{tabular}
    }
\caption{Comparing scores on the \textbf{Validation sets} of different design choices. POS tagging and parsing datasets are averaged. (average over multiple fine-tuning seeds). 
\label{tab:ablation}}
\end{table*}

\section{Impact of Whole-Word Masking}
In Table~\ref{tab:ablation}, we compare models trained using the traditional subword masking with whole-word masking.
Whole-Word Masking positively impacts downstream performances for NLI (although only by 0.5 points of accuracy). To our surprise, this Whole-Word Masking scheme does not benefit much lower level task such as Name Entity Recognition, POS tagging and Dependency Parsing.

\section{Impact of model size}
Table~\ref{tab:ablation} compares models trained with the BASE and LARGE architectures.
These models were trained with the \ccnet corpus (135GB) for practical reasons.
We confirm the positive influence of larger models on the NLI and NER tasks. The LARGE architecture leads to respectively 19.7\% error reduction and 23.7\%.
To our surprise, on POS tagging and dependency parsing, having three time more parameters doesn't lead to a significant  difference compared to the BASE model.
\newcite{tenney-etal-2019-bert} and \newcite{jawahar2019does} have shown that low-level syntactic capabilities are learnt in lower layers of \bert while higher level semantic representations are found in upper layers of \bert.
POS tagging and dependency parsing probably do not benefit from adding more layers as the lower layers of the BASE architecture already capture what is necessary to complete these tasks.

\section{Impact of training dataset}

Table~\ref{tab:ablation} compares models trained on \ccnet and on \oscar.
The major difference between the two datasets is the additional filtering step of \ccnet that favors Wikipedia-Like texts.
The model pretrained on \oscar gets slightly better results on POS tagging and dependency parsing, but gets a larger +1.31 improvement on NER.
The \ccnet model gets better performance on NLI (+0.67).

\section{Impact of number of steps}
\label{sec:nbsteps}

\begin{figure}[t]
\centering
\includegraphics[width=\linewidth]{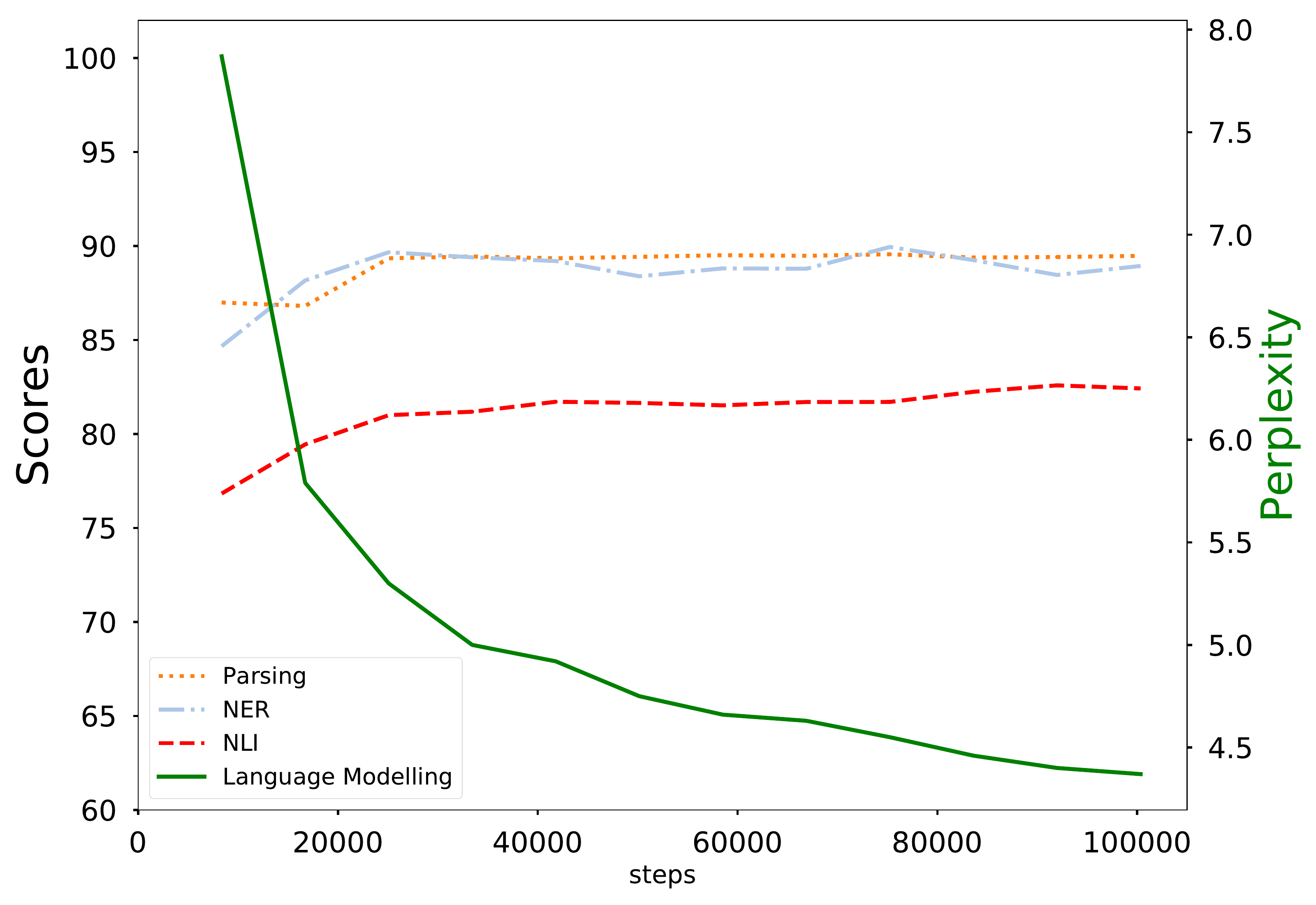}
\caption{Impact of number of pretraining steps on downstream performance for \camembert.}. 
\label{fig:n_steps_impact}
\end{figure}

Figure~\ref{fig:n_steps_impact} displays the evolution of downstream task performance with respect to the number of steps. %
All scores in this section are averages from at least 4 runs with different random seeds. For POS tagging and dependency parsing, we also average the scores on the 4 treebanks.

We evaluate our model at every epoch (1 epoch equals 8360 steps). We report the masked language modelling perplexity along with downstream performances.
Figure~\ref{fig:n_steps_impact}, suggests that the more complex the task the more impactful the number of steps is. We observe an early plateau for dependency parsing and NER at around 22k steps, while for NLI, even if the marginal improvement with regard to pretraining steps becomes smaller, the performance is still slowly increasing at 100k steps.

In Table~\ref{tab:ablation}, we compare two models trained on \ccnet, one for 100k steps and the other for 500k steps to evaluate the influence of the total number of steps.
The model trained for 500k steps does not increase the scores much from just training for 100k steps in POS tagging and parsing.
The increase is slightly higher for XNLI (+0.84).

Those results suggest that low level syntactic representation are captured early in the language model training process while it needs more steps to extract complex semantic information as needed for NLI.

\end{document}